%% file: main.tex
\documentclass[letterpaper, 10 pt, conference]{ieeeconf}  

\IEEEoverridecommandlockouts

\usepackage{amsmath} %

\usepackage{balance}
\usepackage{amssymb}
\usepackage{amsfonts}
\usepackage{enumerate}
\usepackage{tabulary}
\usepackage{multirow}
\usepackage{url}
\usepackage{cite}
\usepackage{lipsum}  
\usepackage{graphicx}
\usepackage{epstopdf}
\usepackage{filecontents}
\usepackage{color}
\usepackage{xcolor}
\usepackage{xxcolor}
\usepackage{setspace}
\usepackage[skins]{tcolorbox}
\usepackage{listings}
\usepackage{lipsum}
\usepackage{booktabs}
\usepackage[lined,boxed,commentsnumbered,linesnumbered]{algorithm2e}
\usepackage{tabularx}
\usepackage{caption}
\usepackage{subcaption}
\usepackage[hidelinks]{hyperref}
\hypersetup{pdftitle={Learned Differentiable Physics for Planning with and Designing
  Deformable-manipulation Tools}}
\definecolor{codegreen}{rgb}{0,0.6,0}
\definecolor{codepurple}{rgb}{0.58,0,0.82}
\definecolor{backcolour}{rgb}{0.95,0.95,0.92}
\lstdefinestyle{buzz}{
    backgroundcolor=\color{black!5},   
    commentstyle=\color{codegreen},
    keywordstyle=\color{blue},
    numberstyle=\tiny\color{black!30},
    stringstyle=\color{codepurple},
    basicstyle=\footnotesize\ttfamily,
    breakatwhitespace=false,         
    breaklines=true,                 
    captionpos=b,                    
    keepspaces=true,                 
    numbers=left,                    
    numbersep=5pt,                  
    showspaces=false,                
    showstringspaces=false,
    showtabs=false,                  
    tabsize=2,
}
\newif\ifanon
\anonfalse         %

\newif\ifarxiv
\arxivtrue         %

\ifanon
  \author{Anonymous Author(s)%
  }
\else
  \author{Nielsen Cugito and Kelsey R. Allen%
  \thanks{Preprint, under review.}%
  \thanks{The authors are with the University of British Columbia and the Vector Institute.}%
  \thanks{Corresponding author: {\tt\small ncugito@cs.ubc.ca}.}%
  }
\fi
\ifanon
  \hypersetup{pdfauthor={Anonymous}}
\else
  \hypersetup{pdfauthor={Nielsen Cugito, Kelsey R. Allen}}
\fi

\begin{document}

\title{\LARGE \bf
    Learned Differentiable Physics for Planning with and Designing Deformable-manipulation Tools
}
\maketitle
\thispagestyle{empty}
\pagestyle{empty}

\begin{figure*}[hbt!]
\includegraphics[width=\linewidth]{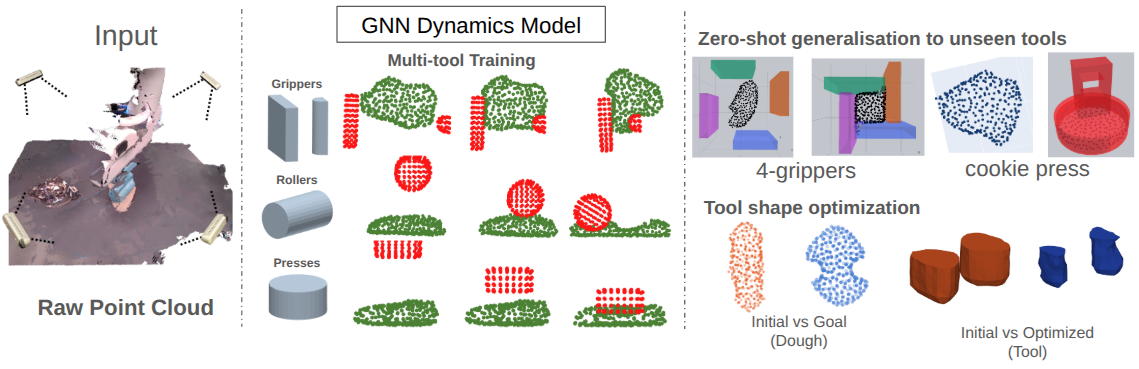}
\caption{\textbf{Method Overview.} Our framework trains a single dynamics model on a variety of tool-object interactions. The model can generalize to tools with shapes that are unseen during training. We further leverage the differentiability of the dynamics model to perform tool shape optimization to achieve target dough shapes.}
\label{Figure1}
\end{figure*}

\begin{abstract}
Model-based planning requires dynamics models that are accurate across a wide range of scene configurations.
For problems like deformable object manipulation, analytic approaches are often too inaccurate, but learning-based techniques often suffer from hallucinations when making predictions outside of their training distributions. 
Here we present an approach for learning dynamics models directly from robot sensor data that is accurate enough for tasks that require designing and using tools for deformable object manipulation.
Our technique uses a graph neural network (GNN) trained on RGB-D observations from a real robot using nine tools spanning three distinct interaction modes.
We introduce a feature parameterization for the GNN which provides rotation and particle density invariance by construction, and show that this is crucial for enabling generalization to new tools and new manipulation trajectories.
Trained on all nine tools, our model outperforms a state-of-the-art machine learning baseline, in both the ``in-distribution" and ``out-of-distribution" setting.
Our model is also more accurate than a differentiable material point method fit to the same data, at an order of magnitude lower cost per planner query.
Finally, its differentiability allows us to perform high degree-of-freedom planning using gradient-based optimization. Given a target dough shape, our model can design a tool to achieve this shape (conditioned on a manipulation trajectory), or come up with a manipulation trajectory for up to 4 grippers to match the target shape.
\ifanon\else
Videos and additional results are available at \url{https://anon-paper-sub.github.io/diffphys-tools/}.
\fi
\end{abstract}

\input{./sections/1-intro.tex}

\input{./sections/2-related.tex}

\input{./sections/4-methods.tex}

\input{./sections/5-results.tex}

\input{./sections/6-conclusions.tex}

\section*{Acknowledgements}
\label{sec:acknowledgements}

\ifanon\else
KRA gratefully acknowledges support from NSERC and CIFAR as both a CIFAR AI
Chair at the Vector Institute, and as a CIFAR Azrieli Global Scholar in the
Learning in Machines and Brains program.
\fi
Generative AI assistants were used to edit and condense the text of this paper
and to help write research and analysis code. All experiments, results and
claims were designed, run and verified by the authors.

\bibliographystyle{./IEEEtranBST/IEEEtran}
\bibliography{./IEEEtranBST/IEEEabrv,./refs}

\ifarxiv
  \clearpage
  \useRomanappendicesfalse  %
  \appendices
  \input{./sections/7-supplementary.tex}
\fi

\end{document}

%% file: sections/1-intro.tex
\section{Introduction}
\label{sec:intro}

Model-based planning plays a crucial role in enabling robots to operate robustly in out-of-distribution (OOD) scenarios. However, it crucially assumes access to an accurate model of the system's dynamics. Recently, there have been many works that employed deep learning techniques on images and point clouds to learn predictive world models that can explicitly reason about future states given an action \cite{huang2026pointworld, wiedemer2025video}. However, the effectiveness of using these learned models for model-based planning is heavily tied to how well they perform outside of the training data. In practice, this is severely understudied. Prior works often use bespoke models engineered for each task \cite{shi2023robocook}, train and evaluate on nearly identical data distributions \cite{whitney2023learning}, or resort to large-scale video modeling \cite{wiedemer2025video} where train and test distributions are rarely cleanly separated.

Analytical models present an alternative, but they are poorly suited for deformable object manipulation. Accurately predicting deformations of soft objects involves high-dimensional, nonlinear dynamics that depend on the object geometry, material properties, and physics parameters. Traditional physics-based simulation methods, such as \cite{jiang2016material, chen2026empm}, require precise system identification and often struggle with real-world variability. Consequently, planning with analytical models becomes intractable in scenarios where generalization is most needed.

In this work, we show that it is possible to learn a generalizable dynamics model directly from RGB-D observations that extrapolates far beyond its training distribution for the task of dough manipulation. From multi-view, real world, RGB-D videos, depth readings are fused together to form a 3D point cloud which represents the dough's state. By taking the current dough state and the next robot action represented as the next frame's tool point cloud position, we train a Graph Neural Networks (GNN)-based dynamics model to predict the next dough state. Our approach trains a single dynamics model on tool-object interactions with different tool shapes, which generalizes to unseen tool geometries during test-time. Remarkably, since our dynamics model is parameterized as a GNN, we can use gradient-based optimization through the dynamics model to optimize new tool shapes or new manipulation trajectories that were never encountered during training to solve a task.

In summary, our main contributions are:
\begin{enumerate}
    \item We present a GNN-based dynamics model that is trained on real robot data. This model takes raw point cloud of the dough and tool point clouds representing actions as inputs, and predict future dough states. A single model is trained jointly on nine different tools, spanning three distinct interaction modes (grasping, rolling, and compressive pushing).
    \item We introduce a $z$-equivariant node and edge parameterization which makes the GNN's predictions exactly equivariant to rotations of the scene about the z-axis, and introduce a split message aggregation, which makes contact messages invariant to the tool surface sampling density. Both improve zero-shot transfer to unseen tools.
    \item We show that our model is differentiable, and its gradients are stable enough to support gradient-based planning to achieve a target dough shape by either (1) optimizing the trajectories of 4 grippers (even though only 2 were ever seen in training), and (2) recovering a real tool's shape from an inflated shape initialization. 
\end{enumerate}

%% file: sections/2-related.tex
\section{Related Work}
\label{sec:related}
\textbf{Deformable Object Simulation.} Modeling accurate deformation of soft objects has been studied for decades. The earliest works that tackled this problem were in the domain of physics-based simulation \cite{o1999graphical, jiang2016material} where constitutive models, together with carefully tuned physical parameters, were employed to model the object's deformation. Recently, methods that use raw sensor readings and perform auto-differentiation with respect to observations have become increasingly popular. Such methods leverage differentiability and physics simulation techniques such as projective dynamics \cite{du2021diffpd, qiao2021differentiable, ma2022risp}, spring-mass systems \cite{zhong2024reconstruction}, and the material point method \cite{chen2026empm, zhang2024physdreamer} to perform system identification of physical parameters to better match observations. Instead of analytically modeling the dynamics of deformable objects, another line of work uses learning-based simulators to model object dynamics \cite{wu2019learning, chen2022comphy, xu2019densephysnet, evans2022context, ma2023learningneuralconstitutivelaws}. These have been applied to learn rigid object dynamics \cite{allen2022learning, whitney2023learning, whitney2024modeling} as well as various types of deformable objects such as granular objects \cite{zhang2024adaptigraph}, plasticine \cite{shi2023robocook, shi2024robocraft}, cloths \cite{pfaff2020learning}, fluids \cite{li2018learning, sanchez2020learning}, and plushies \cite{zhang2024dynamic}. Prior works such as \cite{zhang2024adaptigraph} directly learn object dynamics from real-world videos using appearance-based and geometry losses, which showed generalization to unseen actions. However, most of these learned models were only tested on interactions with a single end-effector. Our work learns a single dynamics model that can generalize to arbitrary end-effectors or tool shapes.

\textbf{Tool Morphology Optimization in Manipulation.} Many works have explored the problem of optimizing designs through differentiable learned simulators. \cite{allen2022physical} uses gradients from learned GNN simulators to perform general-purpose design optimization for surfaces and tools to manipulate fluid flows. In robotics, existing works have also used differentiable simulation to optimize tool morphology for manipulation tasks by representing tools with explicit geometry templates \cite{xu2024dynamics}, rigid links \cite{liu2023learning} and cage-based deformation techniques \cite{li2023learning, guo2024learning, dong2025cagecoopt}. However, these tool morphology parameterizations are often limited in expressiveness and are typically restricted to low-dimensional design spaces. Our method takes inspiration from these prior works, where we leverage a generalizable dynamics model trained on real-world interactions with multiple tools and represent tool shapes with an implicit field. This enables flexible tool morphology deformation in a high-dimensional design space.

%% file: sections/4-methods.tex
\section{Methods}
\label{sec:methods}

\subsection{Dynamics Model}
\label{subsec:dynamics}

\textbf{Problem formulation.} Our goal is to learn a dynamics model $d_{\theta}$ which can map a dough state at time $X_t$ to the dough's state $\Delta$ time-steps into the future: $X_{t+\Delta}$. 
The dough is manipulated by a rigid tool controlled by the robot.
This tool is also represented as a point cloud at a given time-step $T_t$.
Following \cite{shi2023robocook}, actions are represented as the tool's pose trajectory $T_{t:t+\Delta}$. We predict a per-particle displacement for each dough particle $\hat{o}_i = d_\theta(X_{t+\Delta}, T_{t+\Delta})$, such that $\hat{x}_{i,t+\Delta} = x_{i,t} + \hat{o}_i$.   

\textbf{Graph construction.} 
Each step is a graph $G_t = (V_t, E_t)$. Nodes $V_t$ consist of dough particles at time $t$ and the tool particles at $t + \Delta$. Edges $E_t$ consist of dough-dough edges and directed tool$\rightarrow$dough edges. We connect dough-dough particles and tool-dough particles with an edge if the two particles are within the connectivity radius $R$. We additionally apply \emph{normal gating}, keeping a candidate edge only if their normals roughly align. This effectively discards edges passing through the dough interior, which is far in geodesic distance.

\textbf{Node and Edge Features.} We use a $z$-invariant feature parameterization for extracting node and edge features. We exploit the fact that the underlying physical dynamics are $SE(2)$-equivariant: rotating the scene about the vertical $z$-axis rotates the predicted displacements by the exact same angle. We enforce this symmetry by reducing all node and edge vectors to scalar invariants. By incorporating the vertical axis $\hat{z}$ into our dot products (e.g. $d_{rs,z} = d_{rs}{\cdot}\hat{z}$), the features remain sensitive to gravity while remaining z-invariant. 

Under this parameterization, node features are concisely represented by the vertical component of the surface normal $n_{i,z}$, the particle type $\tau_i$ (dough or tool), the floor height $h_i$, and, for dough nodes, the closest tool's signed distance field (SDF) scalar with unit gradient.

To construct the features for an edge $e_{s \rightarrow r}$ connecting a sender node $s$ and a receiver node $r$, we compute scalar invariants from their relative constituent vectors. We build these features using: (1) the relative displacement norm $\|d_{rs}\|$ and its vertical projection $d_{rs,z}$, (2) the difference in particle types $\tau_{rs}$, (3) the pairwise dot products between the normal vectors and the unit displacement $\Phi_{\text{geom}}$ to encode local surface geometry, (4) the vertical projections of the surface normals $n_{r,z}$ and $n_{s,z}$, and (5) the tool velocities at each frame within the step, projected onto the unit displacement, the vertical axis, and the receiver's normal vector $\Phi_{\text{vel}}$. These components provide a complete description of the local interaction while maintaining strict rotational invariance. In summary:
\begin{equation}
  e_{s \rightarrow r} = \big[\, \|d_{rs}\|,\ d_{rs,z},\ \tau_{rs},\ \Phi_{\text{geom}},\ n_{r,z},\ n_{s,z},\ \Phi_{\text{vel}} \,\big]
\label{eq:edge-feat-equivariant}
\end{equation}
By design, applying function transformations on these features, including the message-passing latents, is inherently invariant to $z$-rotations.

\textbf{Processor and decoder.} We follow the standard GNS architecture \cite{sanchez2020learning} where we first embed the node and edge features into latent vectors via an encoder. Then, the processor executes $L$ message-passing steps, allowing local dough-dough and tool-dough interactions to propagate information through the graph. At the end, the decoder predicts per-particle displacement from the dough node latents. 

Since displacement is an equivariant quantity that must rotate \emph{with} the scene, mapping our strictly $z$-invariant latents through a standard MLP would yield $z$-invariant displacements, violating $SE(2)$ equivariance. While prior work often relies on data augmentation to approximate this symmetry \cite{allen2023graph}, which is sample-inefficient, we  guarantee strict equivariance analytically. We train the decoder to output invariant coefficients ($\alpha, \beta$). The final per-particle displacement $\hat{o}_i$ is then computed as a linear combination of these coefficients and an equivariant vector basis, composed of raw node and edge features that naturally change when the scene is rotated around the $z$-axis:
\begin{equation}
  \hat{o}_i = \sum_{k \to i} \alpha_{k \to i}^\top B_{edge}^{(k \to i)} \;+\; \beta_i^\top B_{node}^{(i)}
  \label{eq:equivariant-readout}
\end{equation}
Because a linear combination of equivariant vectors with invariant coefficients is exactly equivariant, this readout strictly preserves $SE(2)$ symmetry. Finally, to maintain this symmetry during denormalization, we apply isotropic motion normalization, where we scale the predicted horizontal displacements uniformly across both $x$ and $y$.

\textbf{Message Aggregation.} Our second architectural deviation modifies how the processor aggregates incoming messages during the node update step. In a standard GNS, a node updates its latent representation by summing all incoming messages into a single pooled vector, regardless of the edge type. However, because tool point clouds can be sampled at arbitrary densities, a sum would cause the magnitude of incoming tool messages to vary drastically depending on the tool's surface sampling density.

To decouple the network's dynamics from the tool's sampling density, we introduce \emph{split aggregation}. We pool the messages from the two distinct edge types separately. For internal dough-dough interactions, we compute a standard sum. For tool-dough interactions, we compute the mean of the incoming messages, which normalizes the signal against the arbitrary number of tool contacts. These separated aggregations are then concatenated with the node's current latent representation $h_i$ and passed through the update network to produce the new latent $h'_i$:
\begin{equation}
  h'_i = \text{MLP}^{\text{processor}} \left( \big[\, h_i,\ m_i^{dough},\ m_i^{tool} \,\big] \right)
\label{eq:split-aggregation}
\end{equation}
The second term corresponds to the standard aggregation of dough-dough edges, while the third term provides a density-invariant aggregation of tool-dough edges. By replacing a single mixed sum with this concatenated structure, the processor can explicitly distinguish between physical dough compression and local tool interactions.

\textbf{Training objective.} We use set-to-set shape terms (similar to \cite{shi2023robocook}):
an equally weighted sum of Chamfer Distance (CD) and Earth Mover's Distance (EMD), and penalize predictions for penetrating the tool and the floor:
\begin{equation}
\mathcal{L}(\mathcal{O}, \hat{\mathcal{O}}) = \tfrac{1}{2}\mathcal{L}_{CD} + \tfrac{1}{2}\mathcal{L}_{EMD} + \lambda_{p}\,\mathcal{L}_{pen} + \lambda_{f}\,\mathcal{L}_{floor}
\label{eq:loss}
\end{equation}
Here $\mathcal{L}_{pen} = \frac{1}{N}\sum_i \max(0,\, -\phi(\hat{x}_i) - \delta)$ is the mean depth by
which predicted particles intrude into the tool beyond a margin $\delta = 1$~mm, $\mathcal{L}_{floor}$ is the same quantity for the floor plane, and $\lambda_{p} = 10$,
$\lambda_{f} = 1$. To stabilize long-horizon prediction the model is trained on $S$ successive steps
rather than one,
\begin{equation}
    \mathcal{L} = \sum_{j=1}^{S}\mathcal{L}(\mathcal{O}_{t + j\Delta}, \hat{\mathcal{O}}_{t + j\Delta}),
\end{equation}
relying on its own prediction $\hat{\mathcal{O}}_{t + (j-1)\Delta}$ for $j > 1$. We use $S = 2$ so a
training sample spans $S\Delta = 6$ frames.

\subsection{Tool shape optimization}
Given a target dough shape, we optimize the tool that produces it by backpropagating through both the
rollout and an implicit representation of the tool's geometry.

\textbf{Tool design parameters.} We feed the initial tool points and normals into a differentiable
Poisson surface reconstruction (DPSR) using the Shape-as-Points \cite{peng2021shape} framework,
obtaining an implicit indicator field for the tool shape. This decouples the resolution used for
simulation from the one used for fabrication: during optimization we re-sample the field's surface at
the density our dynamics model was trained on, and once converged, marching cubes on the same field
at a finer grid gives a smooth, 3D-printable mesh.

\textbf{Design objective and optimizer.} We score a candidate tool by rolling the dynamics model out
over the full action sequence and score with Chamfer distance against the final frame. Since gradient-based optimization scales better to large design spaces than sampling \cite{bharadhwaj2020model}, Adam
updates the tool's surface points and normals through the rollout and the DPSR.

%% file: sections/5-results.tex
\section{Results}
\label{sec:results}
In this section, we show how our proposed graph structured dynamics model enables very strong generalization outside of its training data. We first introduce the dataset, metrics, and baselines, and follow with demonstrating how each proposed component of our proposed dynamics model is necessary for supporting different forms of generalization: accurate predictions for new manipulation trajectories, accurate predictions for new tools, and robustness to different tool sampling densities. We then show how this model enables gradient-based planning to achieve different target dough shapes either by optimizing a tool shape or by planning manipulation trajectories for a large set of grippers. Finally we show that our model also strongly outperforms a state-of-the-art material point method fit to individual tools, and runs an order of magnitude faster.

\subsection{Training data}

We train on the real-robot data from RoboCook \cite{shi2023robocook}, in which a robot arm
manipulates dough with 3D-printed tools and RGB-D cameras record the scene.
One trajectory consists of i.e.\ a single grasp, roll, or press, subsampled to $16$ keyframes,
each storing the fused dough surface as $N = 300$ surface points, and the tool as a point cloud sampled from tool mesh. The nine tools span three interaction modes. The three grippers include Gripper Symmetric Plane (GSP), Gripper Symmetric Rod (GSR),
Gripper Asymmetric (GA). The two rollers include Roller Large (RL), Roller Small (RS). The four presses include 2 presses and 2 punches, each in square and circular
cross-section, Press Square (PRSQ), Press Circle (PRCI), Punch Square (PUSQ), Punch Circle (PUCI). Each tool family variant is shown in Figure~\ref{Figure1}.

\subsection{Metrics}

Every table fixes its test set independently of how the model was trained, so all rows of a table are
scored on the same rollouts. Following \cite{shi2023robocook}, each prediction step advances $\Delta = 3$ keyframes, so a rollout
reaches the last of the $16$ keyframes in $5$ autoregressive steps, each using the previous
step's output. Every table reports the error at the final keyframe rather than an average along
the path.

We evaluate in-distribution and out-of-distribution generalization using
Chamfer Distance (CD) and Earth Mover's Distance (EMD), plus a
signed-distance penetration metric
\begin{equation}
  \mathrm{SDF} = \sum_{i=1}^{N} \max\!\big(0,\, -\phi(x_i) - \delta\big),
  \label{eq:sdf-metric}
\end{equation}
the total depth by which predicted dough particles penetrate the tool,
where $\phi$ is the tool signed-distance field
and $\delta = 1$~mm absorbs signed-distance and tool-pose noise. We sum rather
than average, since averaging over $N$ particles that mostly never touch the
tool would hide how many penetrate. 

\subsection{Baselines}
We primarily compare to RoboCook \cite{shi2023robocook}. In the original paper, \cite{shi2023robocook} trained one model for each tool. We modified their original code to allow training a single model on all tools simultaneously, and report both the single-tool and multi-tool training results here.

\subsection{Generalization experiments}
\textbf{In-distribution experiments. } Every tool is tested under three training regimes -- trained on that tool alone, on its own interaction family, and on all nine tools together.
In each case, the model is scored using that tool's own held-out rollouts.

Our single-tool model matches RoboCook's 9-tool model on both shape metrics, using only $1/9$ the training data (Table~\ref{table_scope_ladder}).
Both formulations improve when training with more tools, but our model performs consistently better. This is most obvious when looking at the penetration metric (SDF), where our model is  approximately $10\times$ better regardless of the amount of training data used.

\newcolumntype{Y}{>{\centering\arraybackslash\hsize=1.6\hsize}X}
\newcolumntype{Z}{>{\centering\arraybackslash\hsize=0.8\hsize}X}

\begin{table}[htb]
    \centering
    \small
    \begin{tabular}{c@{}cccc}
     \toprule
     & Training scope & CD $\downarrow$ & EMD $\downarrow$ & SDF $\downarrow$\\ [0.5ex]
     \hline
     \multirow{3}{0.8cm}{\rotatebox[origin=c]{90}{\shortstack{Robo\\Cook}}}
     &Target tool only & 5.39
     \,{\scriptsize$\pm$\,0.02} & 3.85\,{\scriptsize$\pm$\,0.06} & 1.85\,{\scriptsize$\pm$\,0.41}\\
     &Tool family & 5.35\,{\scriptsize$\pm$\,0.02} & 3.77\,{\scriptsize$\pm$\,0.06} & 0.91\,{\scriptsize$\pm$\,0.14}\\
     &All 9 tools & 5.30\,{\scriptsize$\pm$\,0.02} & 3.68\,{\scriptsize$\pm$\,0.06} & 1.18\,{\scriptsize$\pm$\,0.32}\\
     \hline
     \multirow{3}{0.4cm}{\rotatebox[origin=c]{90}{\textbf{ours}}}
     &Target tool only & 5.28\,{\scriptsize$\pm$\,0.03} & 3.74\,{\scriptsize$\pm$\,0.04} & 0.15\,{\scriptsize$\pm$\,0.12}\\
     &Tool family & 5.25\,{\scriptsize$\pm$\,0.03} & 3.67\,{\scriptsize$\pm$\,0.04} & 0.14\,{\scriptsize$\pm$\,0.11}\\
     &All 9 tools & \textbf{5.20}\,{\scriptsize$\pm$\,0.03} & \textbf{3.58}\,{\scriptsize$\pm$\,0.02} & \textbf{0.11}\,{\scriptsize$\pm$\,0.05}\\
     \bottomrule
    \end{tabular}
    \caption{\textbf{In-distribution effect of multi-tool training.} Error in
    mm, averaged over 30 held-out rollouts per tool. Lower is better; bold is best overall. Our model strongly outperforms RoboCook regardless of training scope.}
    \label{table_scope_ladder}
\end{table}

\textbf{Out-of-distribution experiments.} Our model transfers to tool
shapes it never saw better than RoboCook, on all three metrics (Table~\ref{table_ood}). We trained every model twice, on two complementary sets that partition the nine tools. Each set spans all three interaction modes, so held out means a new shape, not a new interaction. Every tool is held out exactly once. We score each tool on all 211–289 of its trajectories, since none of them are used for training.

RoboCook predicts plausible dough shapes, although with higher tool penetration counts. Adding our penetration losses to that architecture improves the penetration metric (SDF) but leaves the shape metrics unchanged. This suggests that while at least some of the penetration advantage of our model comes from the \emph{objective}, our feature construction (the $z-equivariance$, split message aggregation, and geometry cues) is still important to improve both penetration and shape matching metrics.  Figure~\ref{fig:three_way} shows three held-out
rollouts: on the press, RoboCook's penetration reaches $47.44$~mm at frame $15$ against $1.50$~mm for ours.

\begin{table}[htb]
    \centering
    \small
    \begin{tabular}{cccc}
     \toprule
     Model & CD $\downarrow$ & EMD $\downarrow$ & SDF $\downarrow$\\ [0.5ex]
     \hline
     RoboCook \cite{shi2023robocook} & 5.68\,{\scriptsize$\pm$\,0.01} & 4.26\,{\scriptsize$\pm$\,0.03} & 3.33\,{\scriptsize$\pm$\,0.38}\\
     RoboCook + our losses & 5.66\,{\scriptsize$\pm$\,0.01} & 4.26\,{\scriptsize$\pm$\,0.02} & 1.13\,{\scriptsize$\pm$\,0.27}\\
     \hline
     Ours & \textbf{5.46}\,{\scriptsize$\pm$\,0.01} & \textbf{3.96}\,{\scriptsize$\pm$\,0.03} & \textbf{0.63}\,{\scriptsize$\pm$\,0.34}\\
     \bottomrule
    \end{tabular}
    \caption{\textbf{Zero-shot transfer to unseen tools.} Error in mm, averaged over $\sim250$ held-out rollouts per tool. Intervals are $95\%$ bootstrap over rollouts. Our method
    significantly outperforms RoboCook, especially on the SDF metric, but using
    our losses with RoboCook partially closes the gap.}
    \label{table_ood}
\end{table}

\textbf{Which components support zero-shot transfer.} We retrain our model with one feature construction component removed at a time: the
\emph{heading representation} ($z$-equivariance, the normal dot products,
isotropic motion normalization), \emph{tool discretization} (split
aggregation), and \emph{geometry
cues} (tool SDF feature, normal gating, floor height). 

We focus the ablation experiments on out of distribution generalization for the gripper family (Table~\ref{table_ablation}). For both punches and presses, the ``out-of-distribution" manipulation trajectory is very similar to the training manipulation, differing only in the size or exact shape of the tool, but not how that tool is applied.
In the case of the gripper family, we find a significant effect of both the heading representation and the geometry cues for predicting the correct dough shape, and of the geometry cues for reducing penetration errors. Three seeds of our full model span $0.26$--$0.56$~mm on SDF against $0.07$~mm on CD.

Split aggregation is not expected to have an effect here either: it exists to stop poor scaling with sampling density, but every model here is evaluated at the training density. 

\begin{table}[htb]
    \centering
    \small
    \setlength{\tabcolsep}{3pt}
    \begin{tabular}{lccc}
     \toprule
     Model & CD $\downarrow$ & EMD $\downarrow$ & SDF $\downarrow$\\ [0.5ex]
     \hline
     RoboCook \cite{shi2023robocook} & 6.56\,{\scriptsize$\pm$\,0.03} & 5.45\,{\scriptsize$\pm$\,0.06} & 5.11\,{\scriptsize$\pm$\,0.83}\\
     RoboCook $+$ our losses & 6.52\,{\scriptsize$\pm$\,0.02} & 5.46\,{\scriptsize$\pm$\,0.06} & 1.17\,{\scriptsize$\pm$\,0.55}\\
     \hline
     \textbf{Ours} & \textbf{6.10}\,{\scriptsize$\pm$\,0.04} & \textbf{4.78}\,{\scriptsize$\pm$\,0.05} & \textbf{0.32}\,{\scriptsize$\pm$\,0.07}\\
     \hline
     \hline
     \multicolumn{4}{l}{\textit{Component removals from Ours}}\\
     \hline
     \quad $-$ heading representation & \underline{6.27}\,{\scriptsize$\pm$\,0.05} & \underline{5.20}\,{\scriptsize$\pm$\,0.07} & 0.37\,{\scriptsize$\pm$\,0.05}\\
     \quad $-$ tool discretization & 6.22\,{\scriptsize$\pm$\,0.01} & 4.83\,{\scriptsize$\pm$\,0.02} & 0.37\,{\scriptsize$\pm$\,0.19}\\
     \quad $-$ geometry cues & 6.26\,{\scriptsize$\pm$\,0.04} & 4.93\,{\scriptsize$\pm$\,0.06} & \underline{1.35}\,{\scriptsize$\pm$\,0.75}\\
     \bottomrule
    \end{tabular}
    \caption{\textbf{Ablation study on out-of-distribution generalization.} Error in mm on the three held-out grippers, scored on all
    $837$ of their trajectories; intervals are $95\%$ bootstrap over rollouts. Our full model (bold) achieves the lowest error across all metrics. The bottom section shows the impact of removing individual components, with underlines indicating largest performance drop per metric when a specific component is removed. SDF differences under 0.3mm are within seed variation.}
    \label{table_ablation}
\end{table}

\textbf{Robustness to tool sampling density.} A planner that swaps in a new
tool might use a different sampling density, and the effect of
changing it is substantial (Table~\ref{table_density}). We resample every tool
from $0.25\times$ to $4\times$ the training point cloud density and evaluate with Chamfer Distance. The performance of RoboCook and RoboCook with z-equivariant model degrades significantly at $4\times$ density, since a sum over tool edges grows with the
number of points describing the same geometry. Split aggregation removes that
dependence, barely moving across the full $16\times$ range. 

\begin{table}[htb]
    \centering
    \small
    \begin{tabular}{lccccc}
     \toprule
     Model & $0.25\times$ & $0.5\times$ & $1\times$ & $2\times$ & $4\times$ \\ [0.5ex]
     \hline
     RoboCook \cite{shi2023robocook} & 5.73 & 5.60 & 5.59 & 5.71 & 9.56 \\
     \hline
     $+$ $z$-equivariance & 5.58 & 5.48 & \textbf{5.45} & 5.62 & 13.42 \\
     \hline
     $+$ split aggregation & 5.59 & 5.57 & 5.57 & 5.56 & 5.56 \\
     \hline
     Ours & \textbf{5.50} & \textbf{5.47} & \textbf{5.45} & \textbf{5.44} & \textbf{5.44} \\
     \bottomrule
    \end{tabular}
    \caption{\textbf{Robustness to tool sampling density.}
    Chamfer (mm); $1\times$ is the training density. Bold is best per
    column.}
    \label{table_density}
\end{table}

\begin{figure}[!t]
    \centering
    \begin{subfigure}[b]{\linewidth}
      \centering
      \includegraphics[width=\linewidth]{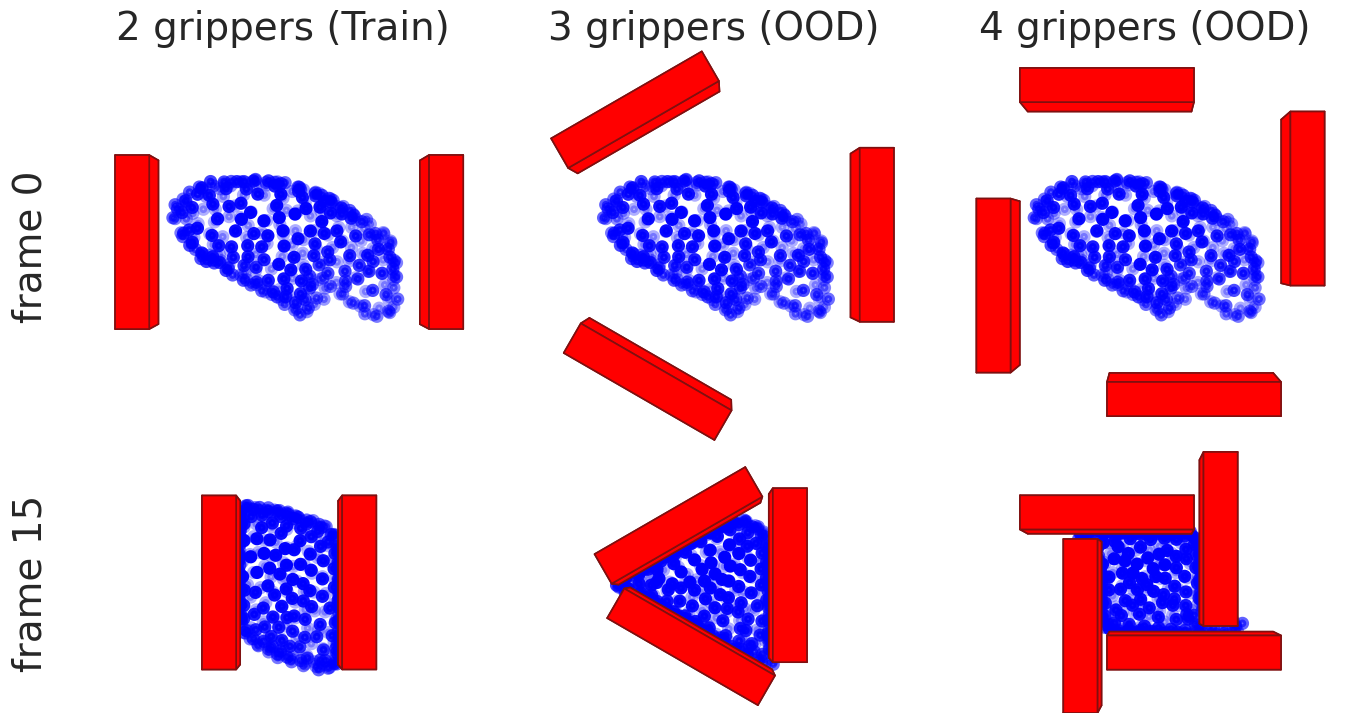}
        \caption{A radial close at two, three and four plate grippers, from above. Two grippers is in-distribution.}
    \end{subfigure}

    \begin{subfigure}[b]{\linewidth}
      \centering
      \includegraphics[width=\linewidth]{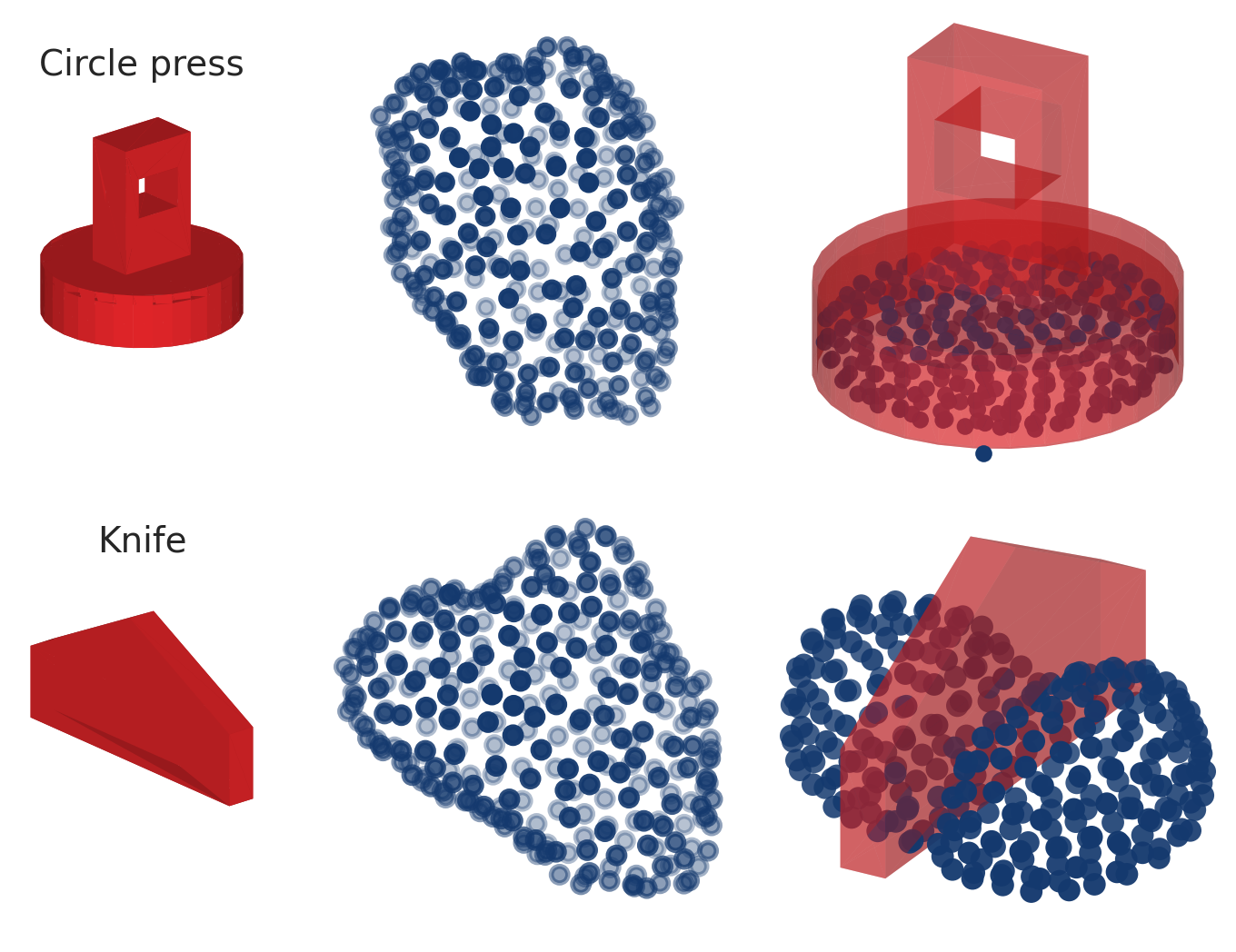}
        \caption{Per row: the tool mesh, the initial dough from above, and the final frame with the tool drawn translucent over the dough it produced.}
    \end{subfigure}
    \caption{\textbf{Qualitative Evaluation.} Tool geometries absent from training. (a)~Plate grippers closing radially, the trained two-gripper pair, then three and four. (b)~A ring press driven to the table, and a knife whose flat face cuts through the dough. Blue is predicted dough, red the tool.}
    \label{fig:cut}
\end{figure}

\textbf{OOD experiments with extra grippers and unseen tool shapes.} In the final generalization experiment, we consider adding more grippers at test time than seen during training, as well as completely new tool geometries. First, we look at three grippers (arranged at $120^\circ$ increments) or four at $90^\circ$, which push dough into the polygon they bound (Fig.~\ref{fig:cut}a). Since we did not collect the initial dataset, we cannot verify that the final dough shapes are precisely what would unfold in the real world. However, qualitative analysis suggests the rollouts are very plausible -- the dough is compressed into the correct polygonal shape, and barely squeezes out of the gaps between the grippers. Whole shapes transfer too (Fig.~\ref{fig:cut}b): a circle press flattens the dough it encloses and a knife successfully separates the dough into two halves.

\subsection{Shape Optimization Experiments}
\label{sec:shapeopt}

\providecommand{\pending}[1]{\textcolor{red}{[#1]}}

We finally ask whether the dynamics model is accurate enough to \emph{design} a
tool, not only to predict with one, by inflating the tool that produced a
ground-truth trajectory and optimizing it back against the goal dough. Goal dough is the ground truth dough at the final frame.

\textbf{The optimized tool produces the target dough. } At $100\%$ inflation scale, rolling out with the inflated tool results in a dough that ends up $40\%$ worse in EMD compared to the original tool. After optimization, all eighteen cells (nine tools, two
inflation scales of $50\%$ and $100\%$, $30$ held-out trajectories each) match the goal dough within
$8.4\%$, and eight match it slightly better than the original tool did.

\textbf{The contact surface converges to the true tool. } On top of matching the target dough, we also want to measure how well we recover to the original tool, especially for particles near the dough interaction region. We take the particles of the true tool that
touch the dough and ask, at each one, how far outside the true surface the candidate tool sits. A tool particle counts as touching if it lies within $6$~mm of the dough the true tool produces at the final frame, just above the $4.3$--$5.5$~mm resampling error of the ground truth dough particles. Inflating by $100\%$ pushes those tool surface particles $5.9$~mm
outside. Optimization recovers $68.2 \pm 7.8\%$ of that distance, against only $37.9 \pm 5.4\%$ on the back of the tool, which has no contact with the dough. The tool recovers nearly twice as well where it contacts the dough, showing that the gradients are highly informative in the contact region.

\begin{figure}[htb]
  \centering
      \includegraphics[width=\columnwidth]{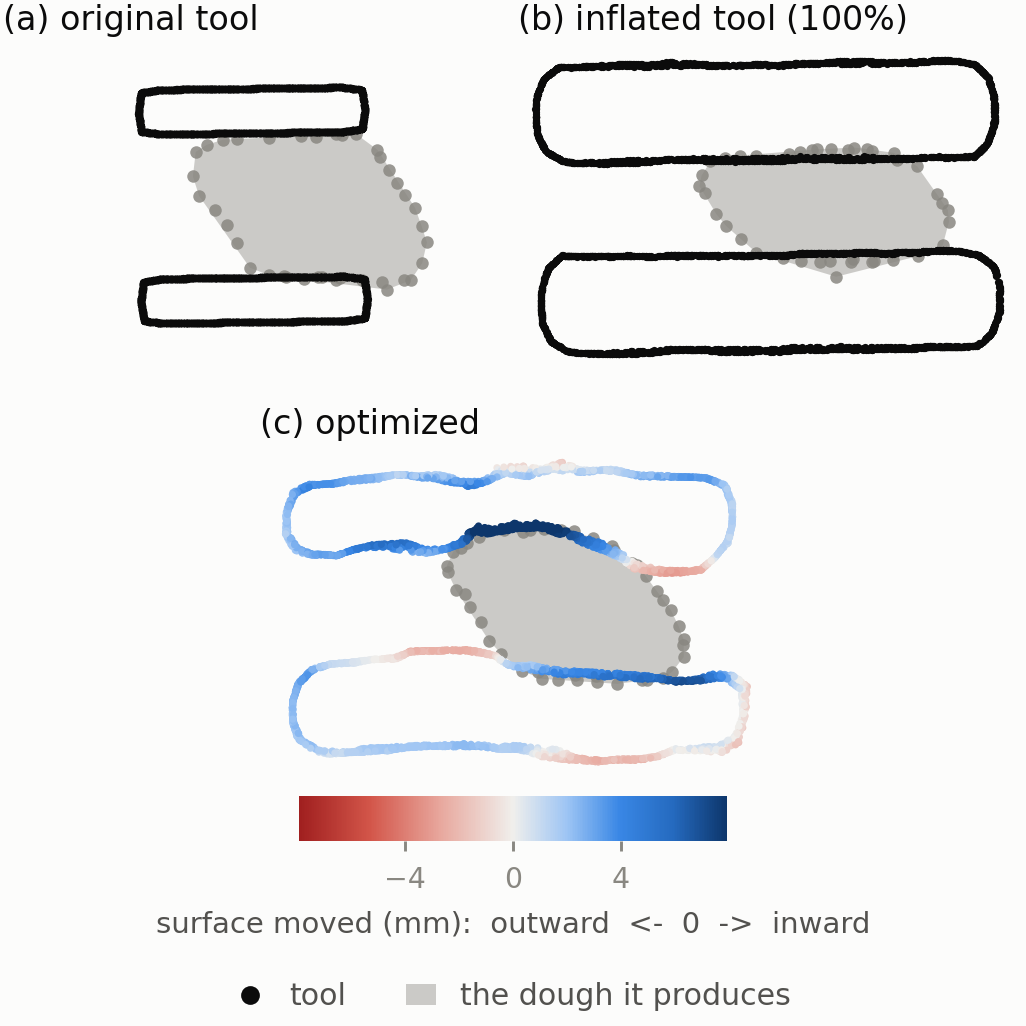}
  \caption{\textbf{What optimization changes, in cross-section.} One GSP
  trajectory at $100\%$ inflation, sliced at the dough's mid-height; each panel
  draws one tool (outline) and the dough it produces (grey), at one scale.
  (a)~Original Tool. (b)~Inflated Tool, dough ends up $40\%$ worse.
  (c)~Optimized Tool, surface coloured by how far each patch moved, blue
  inward (shrinking), red outward (expanding). The tool surface particles touching the dough at the final frame shrink $89\%$ of their inflation ($5.6 \to 0.6$~mm)}
  \label{fig:before_after}
\end{figure}
\begin{figure*}[!ht]
  \centering
  \includegraphics{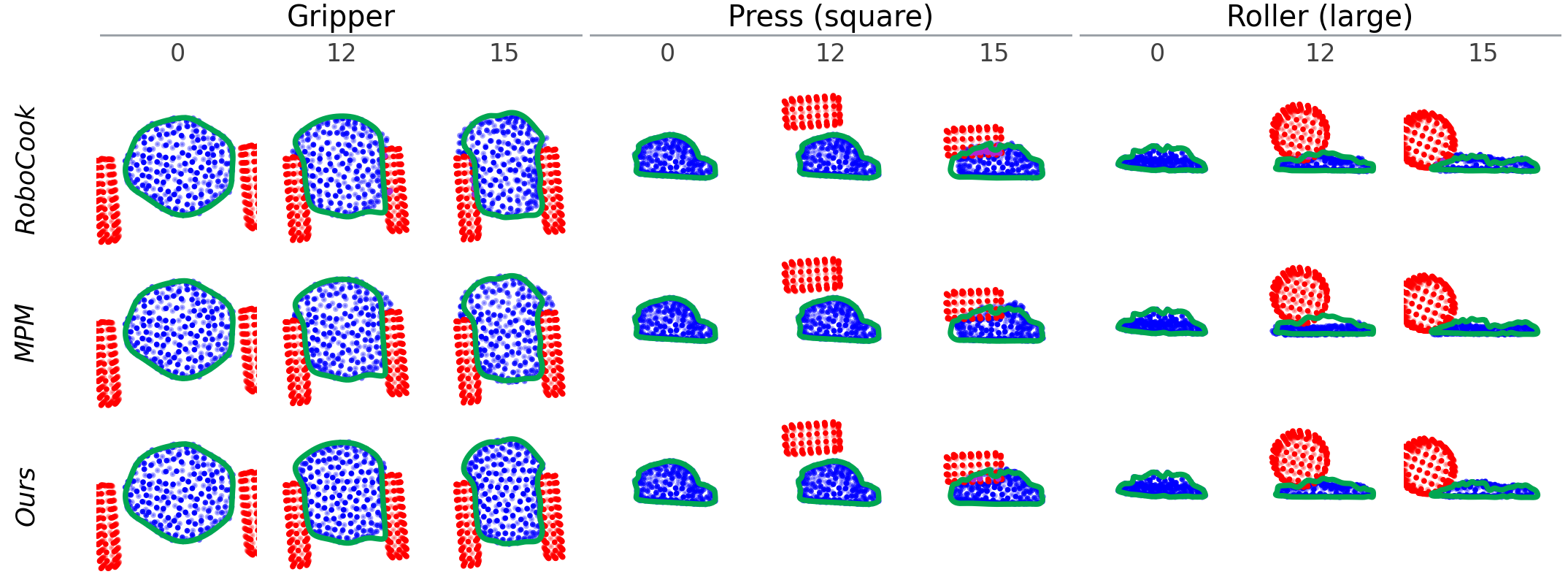}
  \caption{\textbf{Three interaction modes, out of distribution.} One rollout
  per mode, each held out by its own split (Table~\ref{table_ood}): gripper
  (GSP) from above, press (PRSQ) and roller (RL) from the side. Columns are
  frames 0, 12, 15; rows are the models. The press and roller enter after
  frame~0. Blue is the predicted dough, red the tool, green the ground truth outline.}
  \label{fig:three_way}
\end{figure*}

\subsection{Gradient-Based vs. Sampling-Based Planning}

To evaluate the utility of our model for downstream tasks, we design a 48-dimensional planning problem. The task is to recover a scripted four-gripper squeezing action after applying the perturbation. The action space consists of four grippers, each parameterized by an initial $xy$ placement and $xy$ displacements across 5 keyframes. We roll out a reference action through the model to produce a target dough. We then systematically perturb this action by shortening the gripper closures by 10mm, scaling the inward travel of each keyframe and not moving it laterally. Starting from this perturbed state, the optimizer must recover the original action by minimizing a cost function $J(a)$ composed of the  Chamfer distance to the target dough and a heavy penalty for inter-gripper collisions.

We compare two optimization paradigms: gradient descent and the sampling-based Cross-Entropy Method (CEM). Gradient-based optimization leverages the differentiability of our learned simulator, backpropagating the cost function directly through the dynamics model to update the 48-dimensional action vector. In contrast, CEM serves as our derivative-free baseline, maintaining a diagonal Gaussian distribution over the bounded action space. At each iteration, CEM draws a batch of $N = 64$ trajectory samples, evaluates them through the forward model, and refits its distribution to the $E = 8$ lowest-cost actions. By confining both optimizers to an identical bounding box around the perturbed initialization, we rigorously evaluate whether the exact analytical gradients provided by our simulator yield faster, more reliable convergence than high-dimensional stochastic sampling.

\begin{figure}[!t]
  \centering
  \captionsetup[subfigure]{font=footnotesize,justification=centering}
  \begin{subfigure}[b]{0.315\linewidth}
    \includegraphics[width=\linewidth]{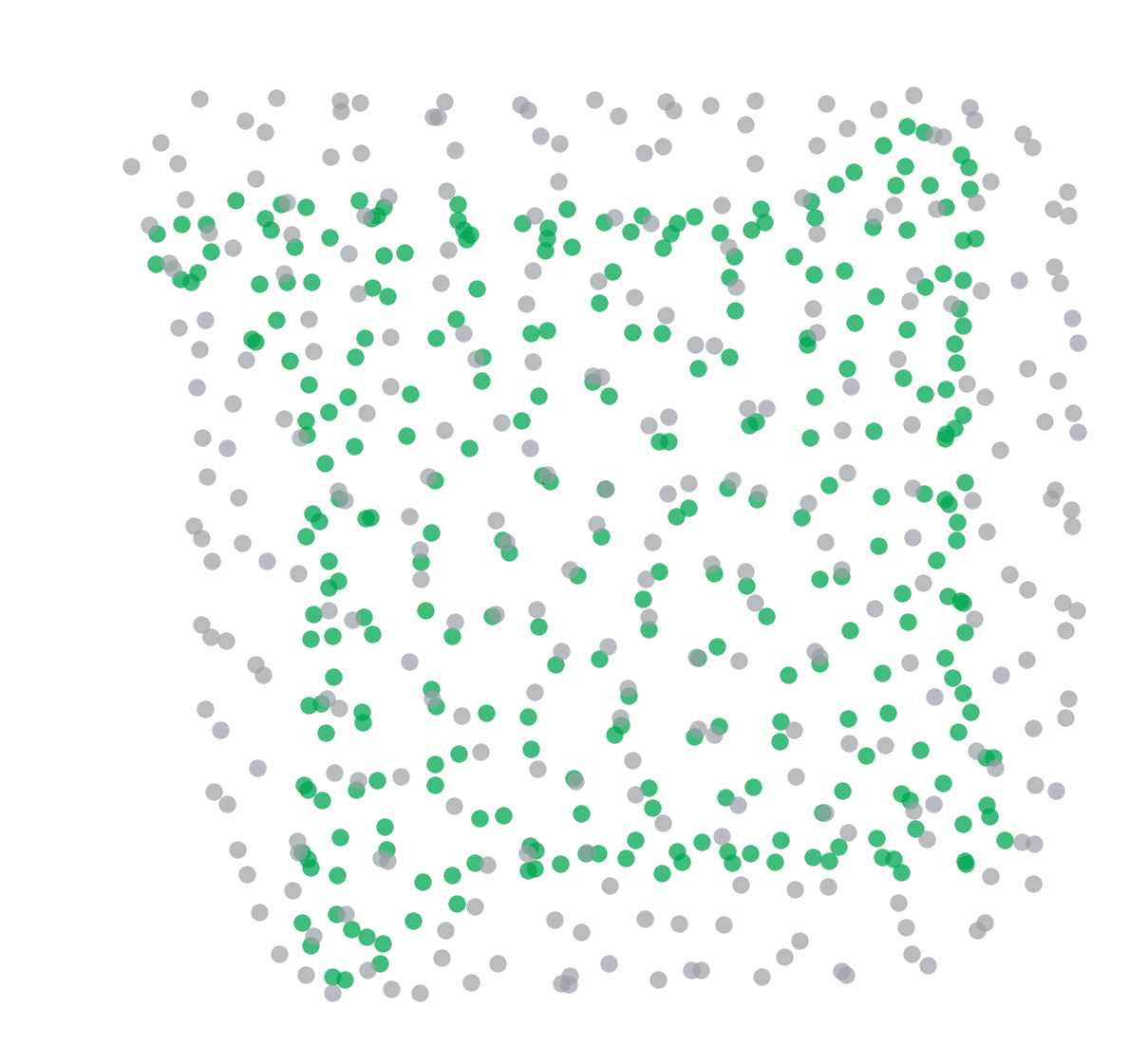}
    \caption{initial vs target}
  \end{subfigure}\hfill
  \begin{subfigure}[b]{0.315\linewidth}
    \includegraphics[width=\linewidth]{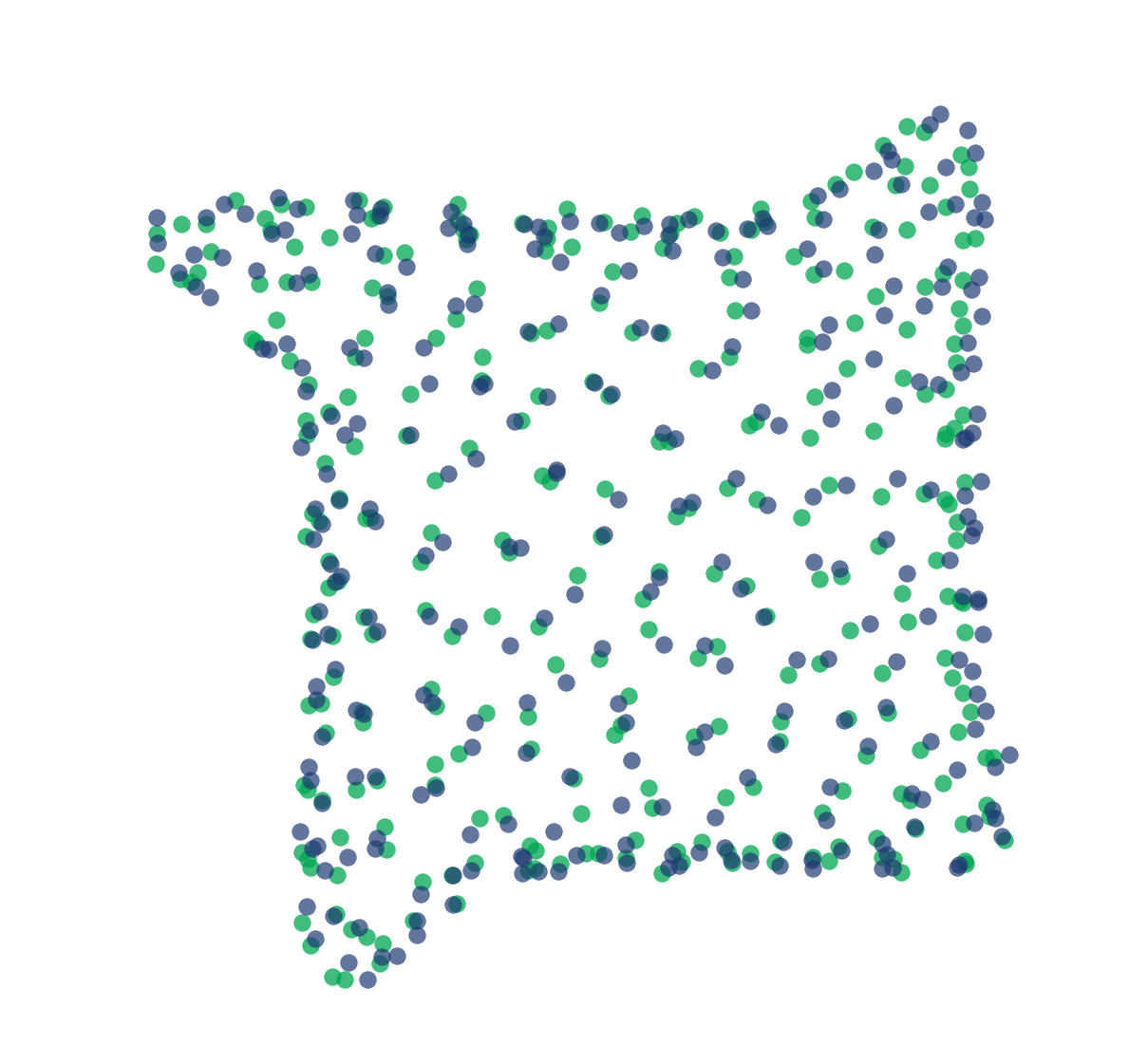}
    \caption{gradient vs target}
  \end{subfigure}\hfill
  \begin{subfigure}[b]{0.315\linewidth}
    \includegraphics[width=\linewidth]{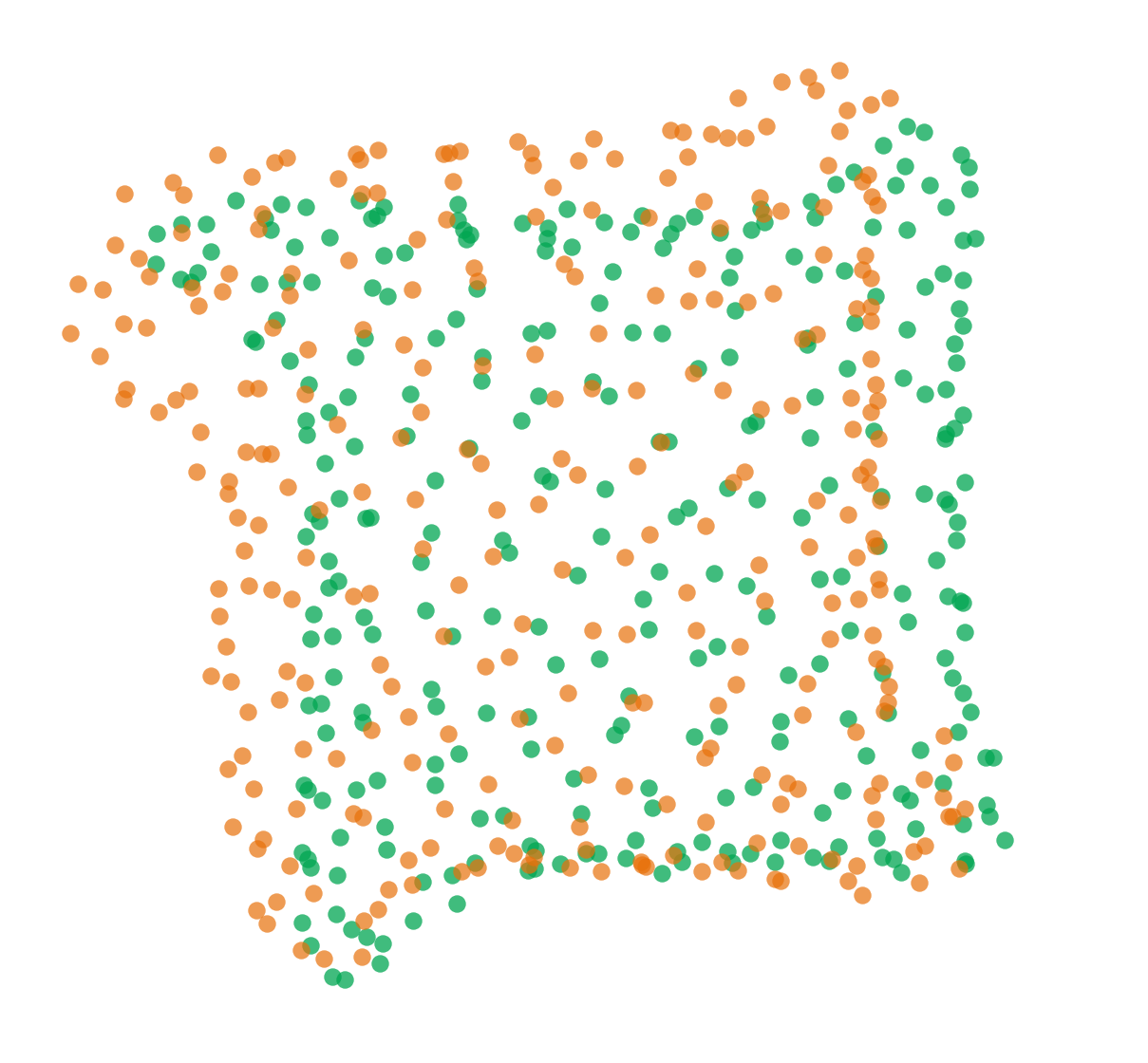}
    \caption{CEM vs target}
  \end{subfigure}

  \vspace{0.4em}
  \begin{subfigure}[b]{0.48\linewidth}
    \includegraphics[width=\linewidth]{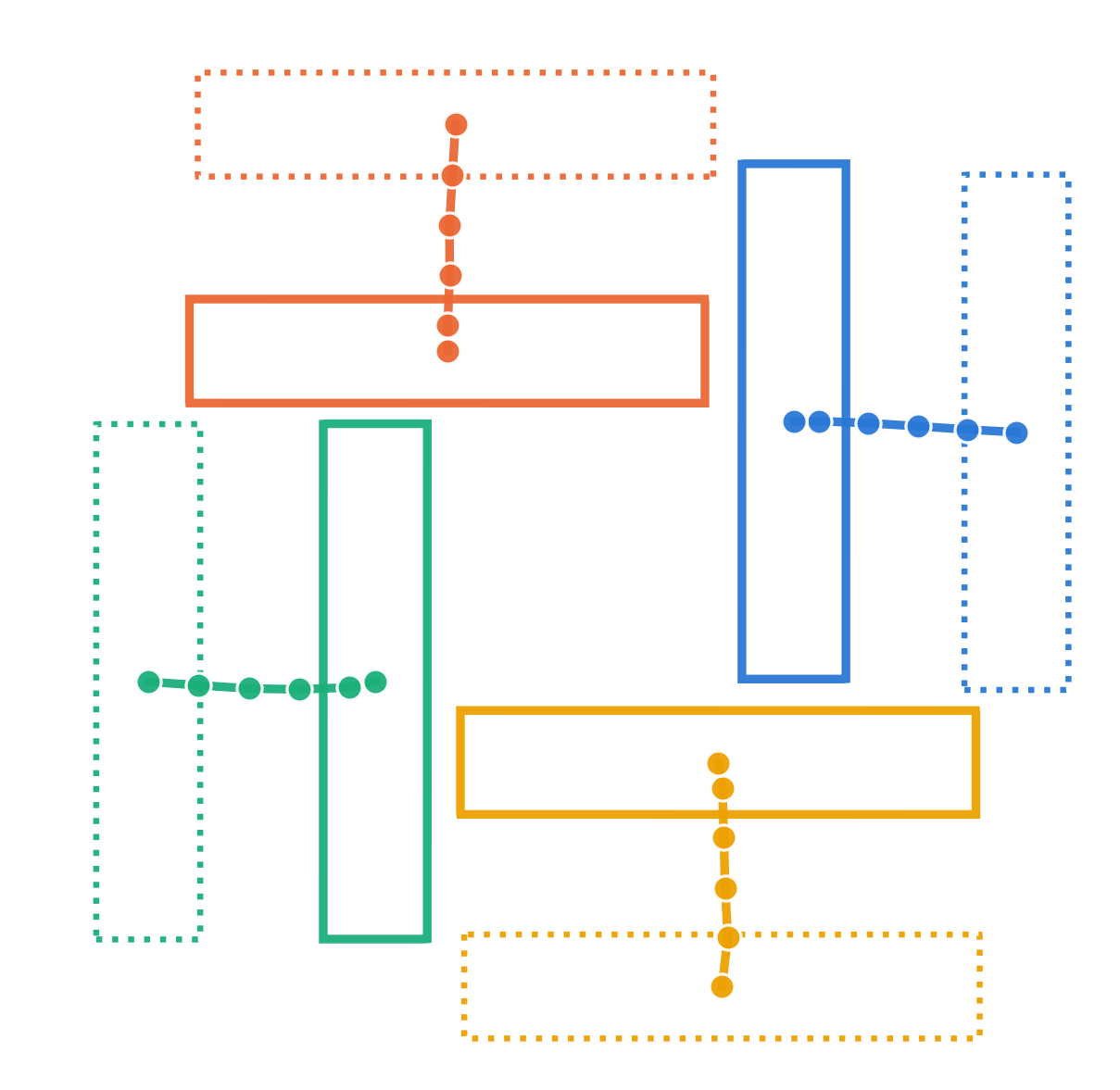}
    \caption{gradient plan}
  \end{subfigure}\hfill
  \begin{subfigure}[b]{0.48\linewidth}
    \includegraphics[width=\linewidth]{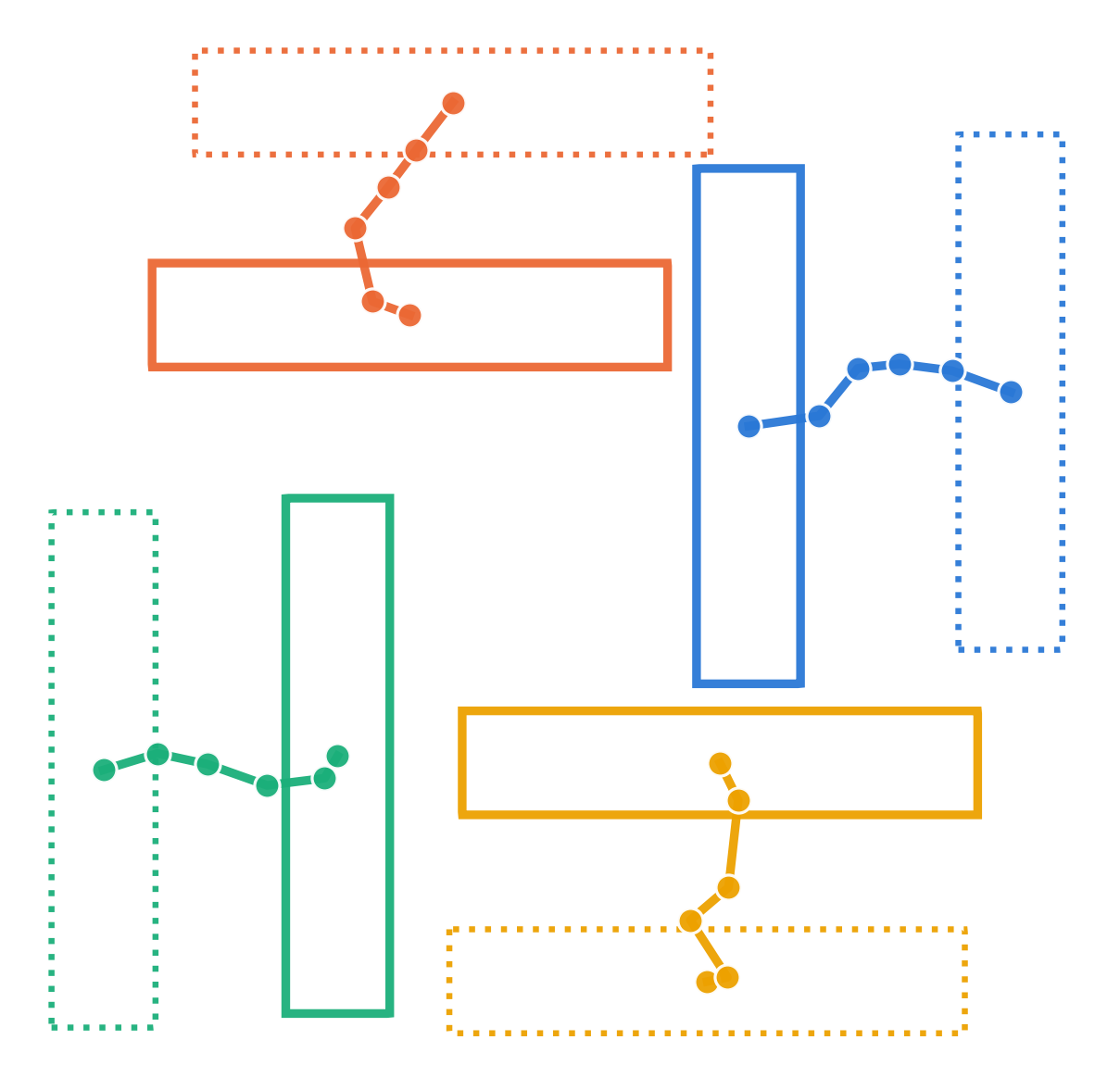}
    \caption{CEM plan}
  \end{subfigure}
  \caption{\textbf{High-DoF action refinement with four grippers}, one held-out
  video seen from above. (a)--(c): the goal dough (green) overlaid with the
  dough from the perturbed initial action, from the gradient-based planning, and from CEM.
  (d), (e): the two plans, one colour per gripper, dotted at the initial pose,
  solid at the final pose.}
  \label{fig:action_refine}
\end{figure}

\textbf{Accuracy and runtime.} Figure~\ref{fig:action_refine} shows the refined plans from both gradient-based and CEM at matched wall-clock. Gradient reaches $1.35$~mm of per-particle RMSE against CEM's $3.37$~mm (20 runs, 10 held-out videos), from a $4.97$~mm perturbation. The gradient converges in $2.7$~s and 78 rollout
equivalents, barely one CEM population, so at matched compute the sampler completes only a couple of
updates. CEM needs 50 updates ($3201$ rollouts, $58$~s), $21\times$ the wall-clock, to
match that accuracy on 9 of 20 runs; on 7 it does not match within 100 updates ($6401$
rollouts, $118$~s).

\subsection{Comparison against EMPM, a Differentiable MPM}

We compare against Embodied MPM (EMPM) \cite{chen2026empm}, a differentiable
Material Point Method for modeling deformable objects from real observations.
MPM is the standard physics-based approach to elastoplastic simulation, and
EMPM identifies its material from recorded interactions. We
follow their formulation, which uses a fixed-corotated elastic model with von Mises
return mapping, the tool entering as a fixed velocity grid boundary with
Coulomb friction, in a differentiable Warp solver, and identify its material
on the same training data our model sees. We compare on \emph{runtime} as a
planning dynamics model and on \emph{prediction accuracy}.

\begin{figure}[!t]
  \centering
  \includegraphics[width=\linewidth]{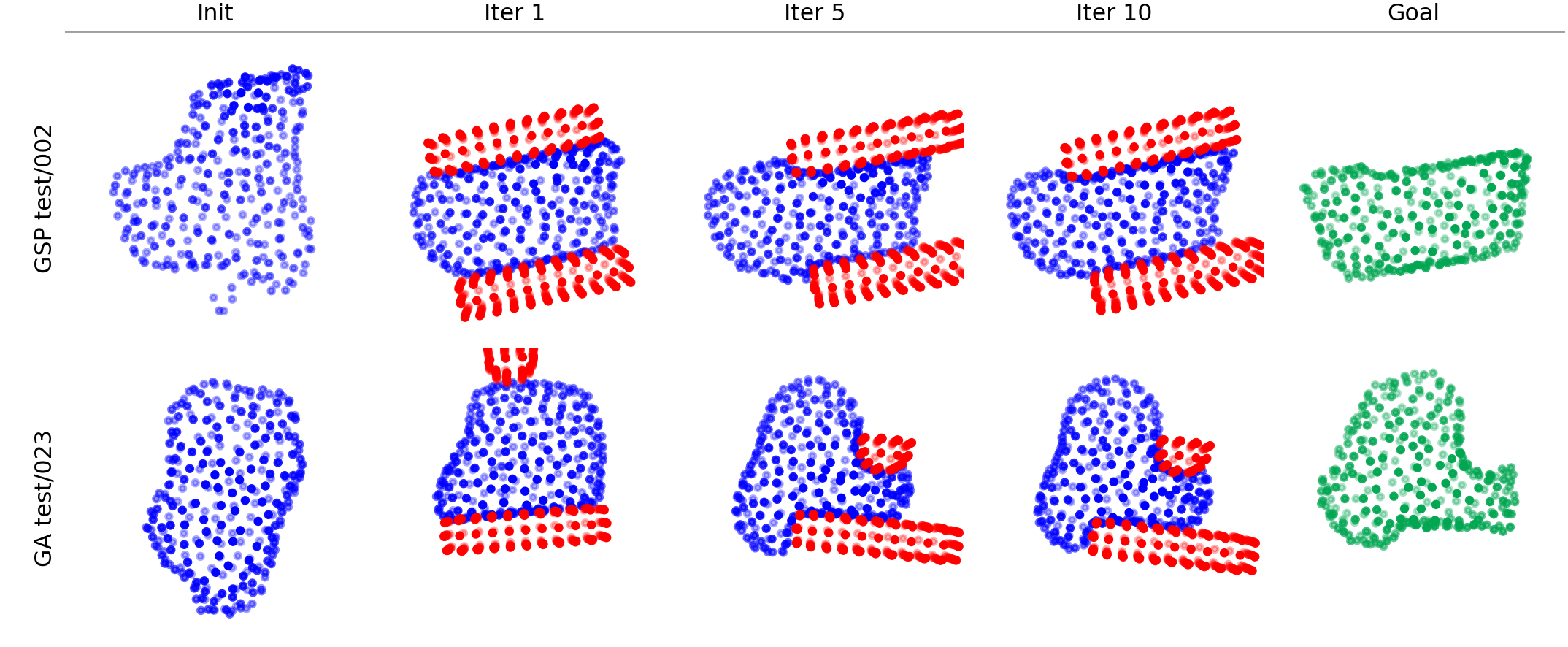}
  \caption{\textbf{CEM planning on the learned model.} Two held-out gripper
  videos seen from above. Iter 1, 5, 10 corresponds to the best candidate of that CEM iteration.}
  \label{fig:cem_iterations}
\end{figure}

\textbf{Runtime as a Planning Dynamics Model.} Our GNN is an order of magnitude
cheaper per planner candidate than EMPM (Table~\ref{table_mpm_runtime}). A
sampling-based planner (CEM) scores a population of candidate trajectories per
decision, so the quantity that matters is the marginal    cost of one more
candidate. Our GNN batches all candidates into a single forward pass; EMPM runs
one sequential rollout per candidate, and within each, a thousand serially
dependent physics substeps in GPU that cannot be batched. A $64$-candidate,
$10$-iteration CEM solve shown in Figure~\ref{fig:cem_iterations}
therefore costs $4.7$~s on our model against $5.5$~min on EMPM.

EMPM's runtime scales linearly with the substep count rather than in the size of the
computation. Shrinking the grid $27\times$ cuts time by only $35\%$, and a
$3.7\times$ increase in particle count leaves it unchanged. The count cannot be
lowered since MPM physics substeps set the physical time per keyframe which sets the tool speed, and the step cannot be enlarged to compensate. The
explicit integrator's stable step is capped by the cell size and the material's
wave speed, where past that cap the simulation diverges.

\begin{table}[htb]
    \centering
    \small
    \begin{tabular}{lr}
     \toprule
     Method & Runtime\\
     \hline
     EMPM & 512.4~ms  \\
     Ours (B=1) & 33.4~ms  \\
     Ours (B=256) & 6.93~ms \\
     \bottomrule
    \end{tabular}
    \caption{\textbf{Per-candidate rollout cost, GNN vs.\ EMPM}, pooled over all nine tools. EMPM runs one sequential rollout for each candidate, while GNN batches candidates in a forward pass which falls to $6.93$~ms at batch size of 256.}
    \label{table_mpm_runtime}
\end{table}

\begin{figure}[!t]
    \centering
    \includegraphics[width=\linewidth]{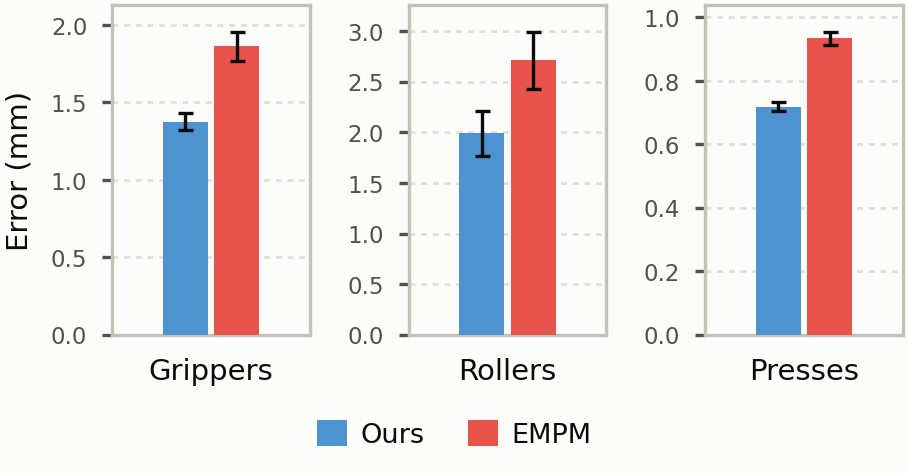}
    \caption{\textbf{Prediction accuracy by tool family, Ours vs.\ EMPM.} Error
    is the point-to-mesh distance: the mean distance from predicted surface point to the ground-truth mesh at the final frame. 30 held-out rollouts per tool; error bars are the standard error over
    rollouts. Ours is the nine-tool model; EMPM runs at 2~mm grid with the identified material.}
    \label{fig_mpm_accuracy}
\end{figure}

\textbf{Prediction Accuracy.} Our GNN is also more accurate, winning both metrics
against every EMPM arm and almost every held-out rollout, even with EMPM's
material fit to our data at 2mm grid resolution
(Figure~\ref{fig_mpm_accuracy}). We
report no penetration for EMPM, whose collider projects particles out of the
tool every substep.

%% file: sections/6-conclusions.tex
\section{Conclusions}
\label{sec:conclusions}

We presented a graph-based dynamics model for deformable manipulation, trained
from multi-view RGB-D observations of a real robot. Building rotation invariance into the graph representation lets a single model span nine tools across three interaction
modes and transfer to geometries it never saw. It improves on a
state-of-the-art learned baseline in and out of distribution, and is more
accurate than a differentiable material point method at an order of magnitude
lower cost per planner query. Differentiability lets the same model design as
well as predict. Its gradients recover a tool that reaches goal dough shape,
and solve action-planning problems in dimensions where sampling-based methods do not scale.

%% file: sections/7-supplementary.tex
\section{QUALITATIVE OUT-OF-DISTRIBUTION ROLLOUTS}
\label{app:ood}

\subsection*{Removing a gripper}
\label{app:onejaw}

\begin{figure*}[p]
  \centering
  \includegraphics[width=\textwidth]{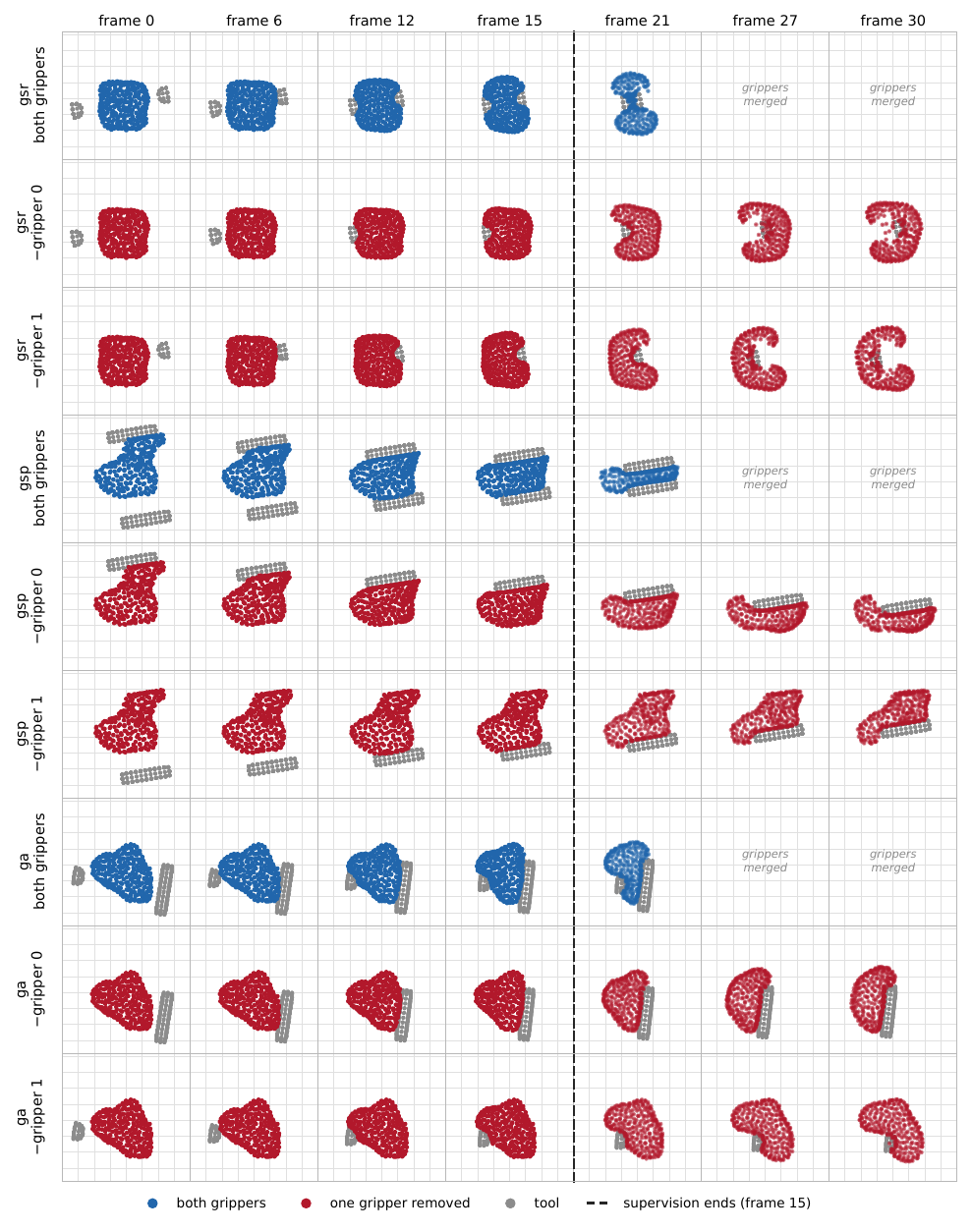}
  \caption{\textbf{Removing a gripper.} Top-down rollouts with both grippers
  (blue) and one gripper removed (red); grey is the tool. Supervision ends at the
  dashed line (frame $15$). Intact rows stop where the two grippers meet.}
  \label{fig:onejaw}
\end{figure*}

Every gripper tool in the training set closes two opposing grippers. To test
whether the model responds to the tool it is actually given, we removed one
gripper and continued the recorded closing motion, so the surviving gripper sweeps
through the dough (Fig.~\ref{fig:onejaw}).

Despite never seeing a single gripper during training, the model predicts
realistic deformation. The surviving gripper pushes the dough ahead of it and cuts
a crescent into it, rather than squeezing it as though the missing gripper were
still there. On GA, the surviving rod slides through the dough while the surviving
plate pushes it.

\subsection*{Extrapolation beyond the supervised horizon}
\label{app:horizon}

The predictions remain stable despite rolling out twice as long as during
training. The rollouts in Fig.~\ref{fig:onejaw} run to frame $30$, twice the
supervised horizon ($10$ autoregressive steps against $5$), and the dough stays a
single coherent body throughout. There is no ground truth past frame $15$, so this
shows stability rather than accuracy.

\section{TEMPORAL RATE CONSISTENCY}
\label{app:rate}

\begin{figure*}[t]
  \centering
  \includegraphics[width=\textwidth]{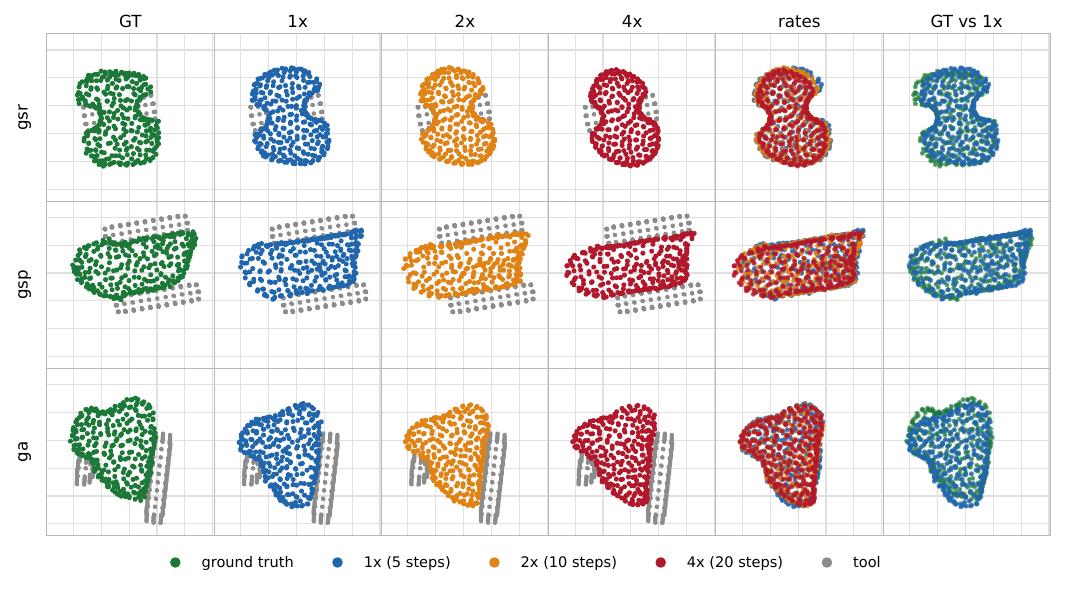}
  \caption{\textbf{Temporal rate consistency.} Final dough from above for one
  tool path rolled out in $5$, $10$ and $20$ steps ($1\times$, $2\times$,
  $4\times$). \emph{rates} overlays the three; \emph{GT vs $1\times$} overlays the
  ground truth and the trained rate.}
  \label{fig:rate}
\end{figure*}

Dough at these speeds is quasi-static (Appendix~\ref{app:mpm}), so traversing the
same tool path more slowly should give the same final shape. We split each step of
the gripper trajectory into $2$ or $4$ smaller ones, keeping the path and the
scored tool poses identical, and rolled out $10$ or $20$ steps instead of $5$.
The predicted shapes stay close across rates (Fig.~\ref{fig:rate}), so the
predicted outcomes stay stable even when the rollout is $2\times$ and $4\times$
longer than the supervised horizon.

\section{DATASET DETAILS}
\label{app:dataset}

We train on the real-robot RoboCook dataset \cite{shi2023robocook}. Each
manipulation is recorded at $5$~Hz over roughly $9$~s by four calibrated RGB-D
cameras and subsampled to $16$ keyframes, uniform in time for the rollers and
uniform in tool displacement for the other tools.

\textbf{Splits.} RoboCook stores each square stamp with its circular counterpart;
we separated them by the circularity of the tool footprint, giving nine tools
rather than seven. We re-partitioned each tool's trajectories under a fixed seed
into $30$ validation, $30$ test and the rest training. This gives $1721$, $270$
and $270$ trajectories in total, and $151$ (RL) to $229$ (GSR) training
trajectories per tool.

\textbf{Training samples.} Each trajectory yields ten chunks of
$1 + \Delta S = 7$ frames. We dropped chunks in which the tool is neither in
contact with the dough nor within one step of it, which removes about half of the
press-style chunks (Table~\ref{table_train_sizes}).

\begin{table}[htb]
    \centering
    \small
    \begin{tabularx}{\linewidth}{|Y|Z|Z|Z|}
     \hline
     Tool & Traj. & Chunks & Kept\\ [0.5ex]
     \hline
     GSR & 229 & 2290 & 100\%\\
     \hline
     GSP & 211 & 2109 & 100\%\\
     \hline
     GA & 217 & 2170 & 100\%\\
     \hline
     RL & 151 & 1510 & 100\%\\
     \hline
     RS & 157 & 1570 & 100\%\\
     \hline
     PRSQ & 176 & 939 & 53\%\\
     \hline
     PRCI & 185 & 911 & 49\%\\
     \hline
     PUSQ & 202 & 996 & 49\%\\
     \hline
     PUCI & 193 & 1021 & 53\%\\
     \hline
     Total & 1721 & 13516 & 79\%\\
     \hline
    \end{tabularx}
    \caption{\textbf{Training-set size per tool.} Trajectories in the
    training split, chunks kept by the contact filter, and the kept fraction of the
    ten chunks per trajectory. Validation and test hold a further $30$ trajectories
    per tool each.}
    \label{table_train_sizes}
\end{table}

\section{BASELINE DETAILS}
\label{app:baselines}

\subsection*{RoboCook}
\label{app:robocook}

We use RoboCook's released code \cite{shi2023robocook}, modified to train one model
on several tools. It is trained on the same splits and schedule as our model
(Appendix~\ref{app:implementation}). \emph{RoboCook $+$ our losses} trains
RoboCook's architecture with our objective, Eq.~(\ref{eq:loss}).

\subsection*{EMPM}
\label{app:mpm}

We follow the formulation of EMPM \cite{chen2026empm}: a fixed-corotated elastic
model with von Mises plasticity, integrated explicitly. The tool enters only as a
boundary condition. Grid nodes inside the tool take its rigid velocity plus
Coulomb friction, and particles found inside are projected back onto its surface
after each substep. This projection enforces near-zero penetration, so we do not
report SDF for EMPM.

\textbf{Simulation settings.} Between keyframes, EMPM integrates $256$ substeps of
$0.05$~ms on a $2$~mm grid, at $5.2$ particles per cell. Each keyframe segment thus lasts $12.8$~ms of simulated time,
far shorter than the real $0.2$ to $0.6$~s. Because the dough is quasi-static,
this barely matters: stretching each segment to $600$~ms changes the final
deformation by only $4.3\%$. A larger $\Delta t$ would make the explicit solver
unstable unless the grid were coarsened in proportion, to cells comparable to the
dough itself, which could no longer resolve its deformation.

\textbf{Material identification.} The simulator's gradients are very noisy, so
we fit the material derivative-free. For each gripper we evaluated a $20$-material grid on every training
rollout, compared materials paired across rollouts, and excluded materials that
fail to integrate. Only yield stress is identified; the other parameters move the
objective about $20\times$ less. One yield stress lies inside all three grippers'
confidence sets.

\textbf{Fitting.} We score $300$ predicted points against the $300$ ground-truth
points under CD and Sinkhorn EMD, since the RoboCook data represents the dough only
as $300$ surface points per keyframe. For EMPM, we initialize the ground-truth
surface points in the particle grid, track them through the simulation, and score
them at the final keyframe against the captured ground truth.

\section{ARCHITECTURE AND TRAINING DETAILS}
\label{app:implementation}

\textbf{Architecture.} We use a connectivity radius $R = 1$~cm, $L = 5$
message-passing rounds and a latent width of $150$ throughout, with layer
normalization after each MLP. Inputs are standardized and the predicted
displacement denormalized with training-set statistics.

\textbf{Message passing.} Each of the $L$ rounds updates the edge messages and
node latents as
\begin{equation}
\begin{split}
  m_k^{\ell} &= \sigma\big(W_{e}\,[\,c_k,\ h_{r_k}^{\ell},\ h_{s_k}^{\ell}\,]\big), \\
  h_i^{\ell+1} &= \sigma\big(W_{n}\,[\,h_i^{0},\ \textstyle\sum_{k \to i} m_k^{\ell}\,] + h_i^{\ell}\big),
\end{split}
\label{eq:message-passing}
\end{equation}
with $\sigma$ a ReLU, $r_k$ and $s_k$ the receiver and sender of edge $k$, and
$k \to i$ the edges received by node $i$. Split aggregation replaces the sum with
the two buckets $m_i^{dough}$ and $m_i^{tool}$, concatenated.

\textbf{Signed-distance fields.} One voxel grid per rigid tool part is built
offline at $1.5$~mm spacing and posed by that part's tracked transform. The
per-part minimum gives the tool's signed distance $\phi$, used as a node feature
and in $\mathcal{L}_{pen}$.

\textbf{Augmentation.} Each training graph draws its tool connectivity radius from
$\mathcal{U}[0.8, 1.2]\,R$ and re-samples the tool cloud at $0.75\times$ to
$2.5\times$ the native density. Neither is applied at evaluation.

\textbf{Optimization.} We train with Adam at learning rate $3 \times 10^{-4}$ and
batch size $16$ for a fixed $110$ epochs, decaying the rate by one decade per $55$
epochs to a floor of $10^{-5}$, with gradient-norm clipping at $1.0$. Batches are
drawn evenly across tools. Checkpoints are selected on full-length validation
rollouts ($30$ per tool), not on the single-step loss.

\section{MULTI-TOOL TRAINING}
\label{app:multitool}

Tool diversity does not substitute for data. We held the training budget fixed at
${\approx}210$ interactions, about $14$~min of manipulation at ${\approx}4$~s
each, and either spent it on one gripper or split it evenly over three ($70$
interactions, ${\approx}4.7$~min per tool). Splitting degrades our
model's CD and EMD on all three grippers (Table~\ref{table_matched_budget}).

Multi-tool training helps the tools with little data. Our single-tool model wins
on exactly the three tools with the most training data and loses on the other six
(Table~\ref{table_specialist_by_data}).

\begin{table}[htb]
    \centering
    \small
    \begin{tabularx}{\linewidth}{|Y|Z|Z|Z|}
     \hline
     Training set & CD $\downarrow$ & EMD $\downarrow$ & SDF $\downarrow$\\ [0.5ex]
     \hline
     \multicolumn{4}{|c|}{\textbf{GSP} (1-tool budget 211)}\\
     \hline
     1 tool & \textbf{5.789} & \textbf{4.397} & 0.027\\
     \hline
     3 tools & 5.951 & 4.572 & \textbf{0.015}\\
     \hline
     \multicolumn{4}{|c|}{\textbf{GSR} (1-tool budget 229)}\\
     \hline
     1 tool & \textbf{5.211} & \textbf{3.513} & \textbf{0.000}\\
     \hline
     3 tools & 5.249 & 3.624 & \textbf{0.000}\\
     \hline
     \multicolumn{4}{|c|}{\textbf{GA} (1-tool budget 217)}\\
     \hline
     1 tool & \textbf{5.933} & \textbf{4.385} & 0.327\\
     \hline
     3 tools & 6.239 & 4.992 & \textbf{0.127}\\
     \hline
    \end{tabularx}
    \caption{\textbf{Tool diversity at a fixed budget,} our model. Each block
    scores one gripper on its $30$ held-out rollouts. The 3-tool model splits ${\approx}210$
    interactions evenly, $70$ per tool, whereas the 1-tool model has ${\approx}210$.
    Error in mm; bold is best per block.}
    \label{table_matched_budget}
\end{table}

\begin{table}[htb]
    \centering
    \small
    \begin{tabularx}{\linewidth}{|Y|Z|Z|Z|}
     \hline
     Tool & Train & $\Delta$CD & $\Delta$EMD\\ [0.5ex]
     \hline
     GSR & 229 & $+0.1$ & $+1.3$\\
     \hline
     GA & 217 & $+2.6$ & $+7.8$\\
     \hline
     GSP & 211 & $+2.5$ & $+1.0$\\
     \hline
     PUSQ & 202 & $-3.5$ & $-7.7$\\
     \hline
     PUCI & 193 & $-3.3$ & $-6.1$\\
     \hline
     PRCI & 185 & $-3.5$ & $-7.8$\\
     \hline
     PRSQ & 176 & $-2.0$ & $-7.3$\\
     \hline
     RS & 157 & $-2.8$ & $-4.6$\\
     \hline
     RL & 151 & $-1.4$ & $-7.3$\\
     \hline
    \end{tabularx}
    \caption{\textbf{Who benefits from multi-tool training.}
    Percentage change from our single-tool to our 9-tool model on each tool's $30$
    held-out rollouts, sorted by training trajectories (``Train''). Negative favours
    the 9-tool model.}
    \label{table_specialist_by_data}
\end{table}

\section{TOOL SHAPE OPTIMIZATION}
\label{app:shaperecovery}

Table~\ref{shape_opt} gives the per-tool dough-space result of
Sec.~\ref{sec:shapeopt}. The contact-surplus recovery of Sec.~\ref{sec:shapeopt}
is stable for probe thresholds of $4$ to $10$~mm ($72.6$ to $69.7\%$ on GSP) and
drops to $56.4\%$ at $15$~mm, as the probe set extends to surface where it does
not contact the dough.

\begin{table}[htb]
    \centering
    \footnotesize
    \begin{tabular}{|l|cc|cc|}
     \hline
      & \multicolumn{2}{c|}{Unopt.\ / GT}
      & \multicolumn{2}{c|}{Optimized / GT} \\
      & $50\%$ & $100\%$ & $50\%$ & $100\%$ \\ [0.3ex]
     \hline
     GA & $111.6\%$ & $126.4\%$ & $96.8\%$ & $104.2\%$ \\
     \cline{1-5}
     GSP & $115.5\%$ & $139.8\%$ & $102.6\%$ & $108.4\%$ \\
     \cline{1-5}
     GSR & $104.8\%$ & $128.0\%$ & $97.4\%$ & $103.1\%$ \\
     \hline
     RS & $110.6\%$ & $116.0\%$ & $90.6\%$ & $90.3\%$ \\
     \cline{1-5}
     RL & $131.3\%$ & $138.0\%$ & $105.6\%$ & $105.3\%$ \\
     \hline
     PRSQ & $115.7\%$ & $139.6\%$ & $93.9\%$ & $94.4\%$ \\
     \cline{1-5}
     PRCI & $116.2\%$ & $135.3\%$ & $101.9\%$ & $102.2\%$ \\
     \hline
     PUSQ & $106.7\%$ & $115.2\%$ & $98.9\%$ & $98.6\%$ \\
     \cline{1-5}
     PUCI & $106.1\%$ & $115.8\%$ & $100.0\%$ & $100.1\%$ \\
     \hline
    \end{tabular}
    \caption{\textbf{Shape optimization in dough space.} Final-frame EMD to the goal
    dough for tools inflated by $50\%$ and $100\%$, before and after optimization,
    as a fraction of the ground-truth tool's, on $30$ held-out rollouts per tool.}
    \label{shape_opt}
\end{table}